%% file: main.tex
\documentclass[a4paper]{article}
\usepackage{graphicx}
\usepackage{twocolceurws}
\usepackage{graphicx}
\usepackage{caption}
\usepackage{lipsum}
\usepackage[utf8]{inputenc}
\usepackage[normalem]{ulem}
\useunder{\uline}{\ul}{}
\usepackage[margin=0.6in]{geometry}
\usepackage[table,xcdraw]{xcolor}
\title{Binary Image Skeletonization Using 2-Stage U-Net}

\author{
Mohamed A. Ghanem \\ oscar@aucegypt.edu
\and
Alaa A. Anani \\  alaa.a.anani@aucegypt.edu
}

\institution{The American University in Cairo \\ Computer Science and Engineering Department}

\begin{document}
\maketitle

\begin{abstract}
 Object Skeletonization is the process of extracting skeletal, line-like representations of shapes. It provides a very useful tool for geometric shape understanding and minimal shape representation. It also has a wide variety of applications, most notably in anatomical research and activity detection. Several mathematical algorithmic approaches have been developed to solve this problem, and some of them have been proven quite robust. However, a lesser amount of attention has been invested into deep learning solutions for it. In this paper, we use a 2-stage variant of the famous U-Net architecture to split the problem space into two sub-problems: shape minimization and corrective skeleton thinning. Our model produces results that are visually much better than the baseline SkelNetOn model. We propose a new metric, M-CCORR, based on normalized correlation coefficients as an alternative to F1 for this challenge as it solves the problem of class imbalance, managing to recognize skeleton similarity without suffering from F1's over-sensitivity to pixel-shifts.
\end{abstract}

\input{introduction}

\input{relatedwork}
\input{dataset}

\input{methodology}
\input{results}
\input{conclusion}

\bibliographystyle{plain} 
\bibliography{ref}

\end{document}

%% file: introduction.tex
\section{Introduction}
Skeletonization is an interesting task because it collapses object shapes into their backbone structure, allowing for constructing a minimal representation of it. Such skeletal representation are useful for many purposes. One simple example of that would be object detection by skeleton matching. Another more involved application is anatomical labelling of different parts of the body such as pulmonary airway trees which requires extracting and identifying different branches of such airways. As such, extracting skeletons can be utilized in shape modelling, compression, and analysis \cite{demir2019skelneton}. One thing to note about skeletons is that they contained condensed local as well as global features. That is, they contain information about the main shape along with peripheral details. 

In spite of the conceptual differences between segmentation and skeletonization, a few famous segmentation models have been observed to perform well when trained for skeletonization. One of these models is the U-Net architecture used for fast segmentation \cite{ronneberger2015u}. U-Net is a fully convolutional neural network that takes its U-shape by virtue of its two main paths: the contraction path, and the expansion path. Passing through the contracting path, the network keeps acquiring features (i.e., information) about \textit{what} the shape its but loses track of \textit{where} they came from. To remedy that, the expansive path tries to reconstruct the image again from the condensed features along with the residual connections with their origin layers. Naturally, the reconstructed image will be a reduced version of the original one, which is nicely aligned with the idea of skeletonization.

%% file: relatedwork.tex
\section{Related Work}
The skeletonization problem has various approaches including heuristical algorithms and using deep neural nets. For the first approach, numerous mathematical algorithms are demonstrated in \cite{SAHA20163}. The problem can also be modelled as a pixel-level binary classification problem known as semantic segmentation. In \cite{demir2019skelneton}, the \textit{SkelNetOn 2019 Challenge} three datasets were proposed to serve as a benchmark for shape understanding approaches, one of which is the \textit{Pixel SkelNetOn}, which poses the challenge in extracting the skeleton pixels from a given binary image. The authors propose a \textit{baseline} model, which is a vanilla \textit{pix2pix} model \cite{isola2018imagetoimage}. They apply distance transform on the input images and \textit{L1} loss scoring an \textit{F1} score of \textbf{0.6244}. In \cite{9025482}, a feature hourglass network (FHN) is used, using a backbone based on \textit{VGGNet} \cite{simonyan2015deep} and side-output integration following it \cite{Liu2017RSRNRS}. They score an \textit{F1} score of \textbf{0.6477}. In \cite{Panichev_2019_CVPR_Workshops}, a modified U-Net based CNN is used alongside the weighted focal loss to overcome the class imbalance problem. Using an ensemble of 8 models they scored an \textit{F1} score of \textbf{0.75} on the validation set. The same idea of binary classification was used by  \cite{xie2015holisticallynested} in their Holistically-Nested Edge Detection (HED) method. The authors use deep supervision on fully connected neural networks and side-output integration. Inspired by the HED method, authors of \cite{nathan2019skeletonnet}, propose their \textit{SkeletonNet} encoder-decoder architecture. Similar to HED, the side layers are fused into final output layers. To improve the HED architecture performance, another type of layers is introduced after the up-sampling stage and their outputs are considered side layers. The loss function used is the sum of the binary cross-entropy and Dice Loss. The best \textit{F1} score of  \textit{SkeletonNet} is \textbf{0.7877}. 

%% file: dataset.tex
\begin{figure*}[!htbp]
\centering
\includegraphics[width=0.9\textwidth]{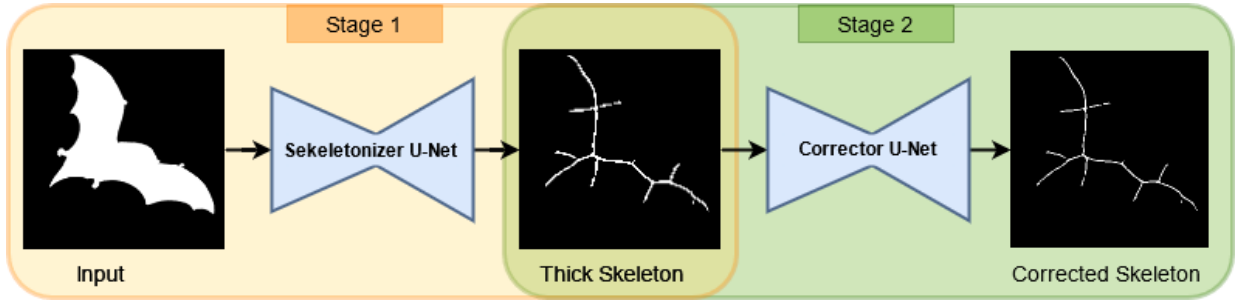}
\caption{High-level view of the 2-stage U-Net architecture composed of the skeletonizer and the corrector models}
\label{fig:archi}
\end{figure*}
\section{Dataset}
In this work, we focus on the Pixel SkelNetOn dataset containing a total of 1,725 binary images on the PNG format of size $256\times 256$ pixels. Objects within each image have been pre-segmented and cleaned. They also contain exactly one object per image. The ground truth was generated algorithmically by \cite{demir2019skelneton} with some manual intervention to remove unnecessary branches. Note that the results stated in this paper are preliminary (on a reserved subset of the training set) as we do not yet have access to the validation set ground truth since it is an ongoing competition.

%% file: methodology.tex
\section{Methodology}
\subsection{Two-stage Problem Split}
In our method, we split the problem space into two consecutive problems handled by two U-Nets that were trained accordingly. The first sub-problem is concerned with condensing or minimizing input shapes which typically results in a thick skeleton-like shape. The second sub-problem expects a bad skeleton and aims to correct as well as thin it. Our training pipeline consists of two such U-Nets in series. Note that the two models are not trained together, that is, the first-stage model is trained on the original data, then the second-stage one is trained on the outputs of stage 1 and the original target skeletons. This is was experimentally observed to produce better results than training both models together.

\subsection{Modified U-Net Architecture}
Both models have the exact same structure and hyper-parameters. A high-level view of our modified U-Net architecture is shown in Figure \ref{fig:archi}. As briefly explained before, a U-Net consists of a contracting path and an expansive one, separated by a bottleneck region. Each contraction (i.e., down-sampling) block contains two convolutional layers with ReLU activation, ending with a $3\times 3$ stride-2 max pooling layer. On the other hand, each expansive (i.e., up-sampling) block contains one deconvolutional layer followed by two convolutional layers and a concatenation layer. The last concatenation layer combines the last output

\subsection{Weighted Categorical Cross-entropy Loss}
Due to the fact that background pixels heavily outnumber skeleton pixels in target images, mistakes in each of them cannot be weighted equally. This is a classic issue of class imbalance which we address in our loss by giving higher weight to misclassifications on the positive class (i.e., skeleton). This incentivizes the model to be more careful with the skeletal region. On that basis, our loss function is a weighted categorical cross-entropy (CCE) computed as:
\begin{center}
    $\mathcal{L}(p_t, p_{r}) = 1 - w*p_t*p_{r}$
\end{center}

where $w$ is the class weights ($w_0 \ll w_1$), $p_t$ is the true class (0 for background, 1 for skeleton), and $p_{r}$ is the predicted probability.

%% file: results.tex
\begin{figure*}[!htbp]
\centering
\includegraphics[width=0.9\textwidth]{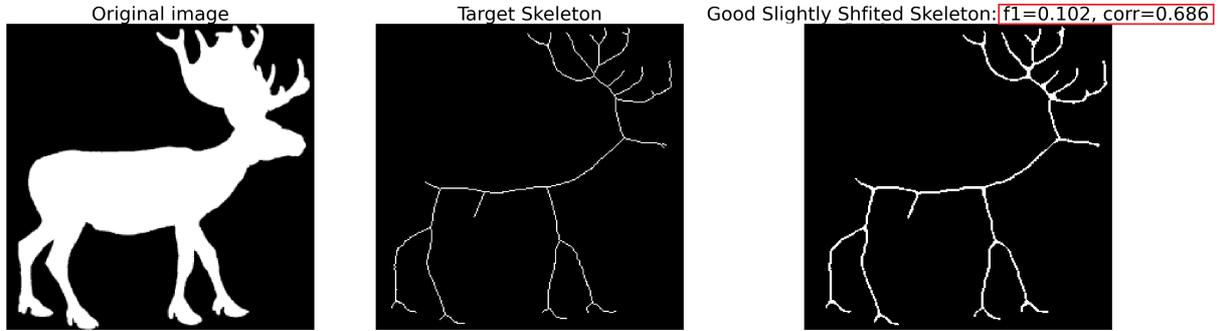}
\caption{Output example for the \textit{f1} and correlation scores (M-CCORR=corr) of a good, but manually shifted skeleton output. Though the skeletons look alike, the \textit{F1} score is too low while M-CCORR shows a reasonable number}
\label{f1sucks}
\end{figure*}
\section{Results}
\subsection{Experimental Setup}
We have experimented with multiple U-Net based architectures using different losses and weights. Although there are differences among the architectures, there are some parameters that are fixed across all experiments. These parameters are shown in Table \ref{table:params}.
\captionsetup{belowskip=-15pt}
\begin{center}
\begin{table}[h!]
\centering
 \begin{tabular}{l|l} 
 \hline
 \rowcolor[HTML]{ACDAFC} 
 \textbf{Parameter} & \textbf{Value} \\ [0.5ex] 
 \hline
 Optimizer & Adam \\ 
 \hline
 Learning Rate & 0.001 \\ 
 \hline
 $\beta_1$ & 0.9 \\ 
 $\beta_2$ & 0.999 \\ 
 \hline
 Batch Size & 32 \\ 
 \hline
 loss & Weighted CCE \\
 \hline
 Loss weights & [1, 25] \\
 \hline
 Metrics & [F1, M-CCORR] \\
 \hline
\end{tabular}
\caption{Hyper-parameters across architectural experiments}
\label{table:params}
\end{table}
\end{center}

\subsection{New Metric Proposal}
Though the \textit{F1} score is used widely on this dataset as an evaluation metric \cite{demir2019skelneton, 9025482, Panichev_2019_CVPR_Workshops, nathan2019skeletonnet}, one issue that arises with it is that it is very sensitive to small pixel-offset errors. If the skeleton image and the ground truth are partially translated (even by few pixels), the \textit{F1} score decreases dramatically (check Figure \ref{f1sucks} for illustration). This pixel-offset sensitivity poorly evaluates good-looking skeletons though they logically match the ground truth. Hence, we propose the usage of template matching as a new translation-aware metric for this problem. We have developed a variant of the normalized cross-correlation coefficient which we call the Matching Cross-correlation (M-CCORR) as defined in the following equation:
\[\textit{M-CCORR}(y_t, y_p)= \frac{max(CCORR(y_t, y_p))}{
log_2(\mathcal{D}(y_t, y_p)+2)}\]
where $y_t$ and $y_p$ are the ground truth and prediction output images respectively and $\mathcal{D}$ is a function that return the distance between the bounding box center of the ground truth skeleton and that calculated on the predicted skeleton. This denominator ensures that the \textit{Matching} CCORR-based metric we are proposing is not only aware of pixel-offsets but also of the logical resemblance between the target and prediction.

\subsection{Experiments}
We have experimented with multiple stages of our U-Net based architectures, through varying the number of corrector after the skeletonizer models as shown in Figre \ref{fig:archi}. The architectures are the following:
\begin{itemize}
\item \textbf{One-Stage (Skeletonizer)}:
The entire pipeline consists of a signle model that gets trained to perform the whole skeletonization process.
\item \textbf{Two-Stage (Skeletonizer + Corrector)}:
The training pipeline consists of two models: a skeletonizer that produces rough skeletons and a corrector which enhances rough skeletons. 
\item \textbf{Three-Stage (Skeletonizer + 2 Correctors)}:
The training pipeline consists of three models: a skeletonizer and two correctors in series. This adds an extra enhancement layer on its 2-stage predecessor. 
\end{itemize}
The following Table \ref{table:overall} shows our $F1$ and M-CCORR scores corresponding to every $n$-stage experiment:
\begin{center}
\begin{table}[h!]
\centering
 \begin{tabular}{l|l|l} 
 \hline
 \rowcolor[HTML]{ACDAFC} 
 \textbf{Model} & \textbf{F1 Score} & \textbf{M-CCORR} \\ [0.5ex] 
 \hline
One-Stage & 0.4866 & 0.5604\\ 
 \hline
Two-Stage & \textbf{0.5968} & \textbf{0.6399} \\ 
 \hline
Three-Stage & 0.5802 & 0.6271 \\
\hline

\end{tabular}

\caption{\textit{F1} and M-CCORR scores corresponding to different number of stages of the $U$-$Net$ like model}
\label{table:overall}
\end{table}
\end{center}

The two-stage model outperforms both one-stage and three-stage models in both \textit{F1} and M-CCORR scores. It has an \textit{F1} score of \textbf{0.5968} and an M-CCORR score of \textbf{0.6399} on the validation set. A single skeletonizer produces thick skeletons causing many false positives reducing the \textit{F1} score, which is fine-tuned by the corrector following it in the two-stage model. Adding more than one corrector (i.e, the three-stage model) does not further improve the results. 

In addition to the above setups, we have explored multiple other modifications that did not further improve the \textit{F1} and M-CCORR scores. These include (but are not limited to): L1, hinge and focal losses, adding a dilation layer before the output, and applying a distance transform on the input images.

%% file: conclusion.tex
\section{Conclusion}
In summary, our preliminary results have shown that splitting the problem of skeletonization into two stage does indeed bring significant improvement of accuracy. We have also shown that there is a limit to the split-level beyond which performance will not improve. The literature has several solutions that perform very well in one stage, so incorporating them into a similar 2-stage workflow can boost their accuracy even further. Last but not least, we have demonstrated how sensitivity to translation in the F1 score undermines its representativeness of skeleton quality - proposing normalized correlation coefficients in its stead.

%% file: main.bbl
\begin{thebibliography}{10}

\bibitem{demir2019skelneton}
Ilke Demir, Camilla Hahn, Kathryn Leonard, Geraldine Morin, Dana Rahbani,
  Athina Panotopoulou, Amelie Fondevilla, Elena Balashova, Bastien Durix, and
  Adam Kortylewski.
\newblock Skelneton 2019: Dataset and challenge on deep learning for geometric
  shape understanding.
\newblock In {\em Proceedings of the IEEE/CVF Conference on Computer Vision and
  Pattern Recognition Workshops}, pages 0--0, 2019.

\bibitem{isola2018imagetoimage}
Phillip Isola, Jun-Yan Zhu, Tinghui Zhou, and Alexei~A. Efros.
\newblock Image-to-image translation with conditional adversarial networks,
  2018.

\bibitem{9025482}
Nan Jiang, Yifei Zhang, Dezhao Luo, Chang Liu, Yu~Zhou, and Zhenjun Han.
\newblock Feature hourglass network for skeleton detection.
\newblock In {\em 2019 IEEE/CVF Conference on Computer Vision and Pattern
  Recognition Workshops (CVPRW)}, pages 1172--1176, 2019.

\bibitem{Liu2017RSRNRS}
Chang Liu, Wei Ke, Jianbin Jiao, and Qixiang Ye.
\newblock Rsrn: Rich side-output residual network for medial axis detection.
\newblock {\em 2017 IEEE International Conference on Computer Vision Workshops
  (ICCVW)}, pages 1739--1743, 2017.

\bibitem{nathan2019skeletonnet}
Sabari Nathan and Priya Kansal.
\newblock Skeletonnet: Shape pixel to skeleton pixel, 2019.

\bibitem{Panichev_2019_CVPR_Workshops}
Oleg Panichev and Alona Voloshyna.
\newblock U-net based convolutional neural network for skeleton extraction.
\newblock In {\em Proceedings of the IEEE/CVF Conference on Computer Vision and
  Pattern Recognition (CVPR) Workshops}, June 2019.

\bibitem{ronneberger2015u}
Olaf Ronneberger, Philipp Fischer, and Thomas Brox.
\newblock U-net: Convolutional networks for biomedical image segmentation.
\newblock In {\em International Conference on Medical image computing and
  computer-assisted intervention}, pages 234--241. Springer, 2015.

\bibitem{SAHA20163}
Punam~K. Saha, Gunilla Borgefors, and Gabriella {Sanniti di Baja}.
\newblock A survey on skeletonization algorithms and their applications.
\newblock {\em Pattern Recognition Letters}, 76:3--12, 2016.
\newblock Special Issue on Skeletonization and its Application.

\bibitem{simonyan2015deep}
Karen Simonyan and Andrew Zisserman.
\newblock Very deep convolutional networks for large-scale image recognition,
  2015.

\bibitem{xie2015holisticallynested}
Saining Xie and Zhuowen Tu.
\newblock Holistically-nested edge detection, 2015.

\end{thebibliography}
